\documentclass[letterpaper]{article}
\usepackage[preprint]{aaai2027}

\usepackage[hyphens]{url}
\usepackage{graphicx}
\usepackage{natbib}
\usepackage{caption}
\usepackage{algorithm}
\usepackage{algorithmic}

\usepackage{newfloat}
\usepackage{listings}
\DeclareCaptionStyle{ruled}{labelfont=normalfont,labelsep=colon,strut=off}
\floatstyle{ruled}
\newfloat{listing}{tb}{lst}{}
\floatname{listing}{Listing}

\usepackage{booktabs}

\usepackage[table]{xcolor}
\newcommand{\deemph}{\color{black!55}}

\usepackage{amssymb}
\usepackage{pdfpages}

\definecolor{aaailink}{RGB}{0,71,133}
\definecolor{aaaicite}{RGB}{0,102,84}
\definecolor{aaaiurl}{RGB}{135,45,110}
\usepackage[
  colorlinks=true,
  linkcolor=aaailink,
  citecolor=aaaicite,
  urlcolor=aaaiurl,
  linktocpage=true,
  breaklinks=true,
  bookmarksnumbered=true,
  pdfstartview=FitH
]{hyperref}

\hypersetup{
  pdftitle={TraceViT: Grounded Trace Supervision for Visual Abstract Reasoning},
  pdfauthor={Binnan Liu, Yechi Ma, Tian Xie, Wei Hua},
  pdfsubject={Visual abstract reasoning; ARC-AGI; looped transformers},
  pdfkeywords={abstract reasoning, ARC-AGI, process supervision, vision transformer}
}

\newcommand{\availabilitystatement}{%
Code and data will be available at
\url{https://github.com/LiuBinnan/TraceViT}.}

\title{TraceViT: Grounded Trace Supervision for Visual Abstract Reasoning}

\author{
    Binnan Liu\textsuperscript{\rm 1,2},
    Yechi Ma\textsuperscript{\rm 1,2},
    Tian Xie\textsuperscript{\rm 1},
    Wei Hua\textsuperscript{\rm 1,2}\corresponding
}

\affiliations{
    \textsuperscript{\rm 1}Zhejiang University\\
    \textsuperscript{\rm 2}Zhejiang Lab\\
    binnanliu@zju.edu.cn, huawei@zhejianglab.com
}

\begin{document}

\maketitle


\begin{abstract}
The Abstraction and Reasoning Corpus (ARC) tests whether a model can
infer an unseen transformation from a few input--output examples and
apply it to a new grid. Looped visual reasoners refine
predictions over multiple iterations, but conventional training
constrains only the final output, leaving intermediate refinements unconstrained. We propose that these refinements should instead follow the
transformation step by step. We introduce TraceViT, a looped visual reasoner trained with
semantically monotonic transformation chains. We obtain these chains by
rewriting and verifying programmatic task implementations, decomposing
each solution into intermediate grid states. Each iteration is grounded by a task reference derived from the
few-shot demonstrations and an object workspace representing the
current grid state. Because these chains may differ in length from the loop, soft trace
alignment enforces only their ordering, letting the model allocate
iterations freely. TraceViT achieves 67.8\%
pass@2 on ARC-AGI-1 and 24.3\% on ARC-AGI-2. Controlled ablations on ARC-AGI-1 show that trace supervision
becomes beneficial only when paired with grounding.
\availabilitystatement
\end{abstract}

\section{Introduction}

\begin{figure}[t]
\centering
\includegraphics[width=\columnwidth]{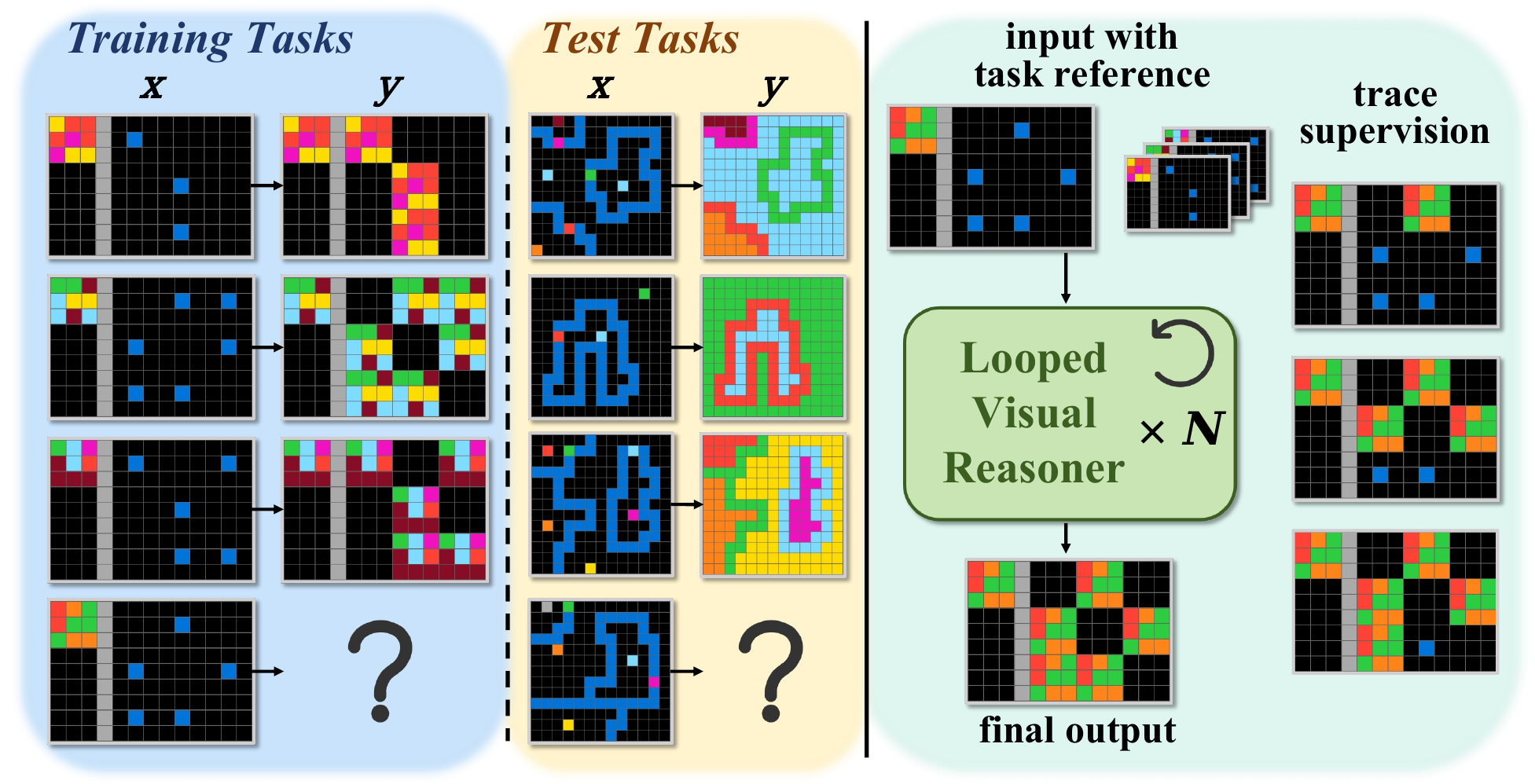}
\caption{\textbf{Grounded trace supervision.} \textbf{Left:} an ARC
task provides demonstrations and a test input. \textbf{Right:}
TraceViT iterates a shared visual core, decoding a grid at every
step. During training, intermediate predictions are aligned with
transformation chain milestones. See Fig.~\ref{fig:method} for
the full pipeline.}
\label{fig:teaser}
\end{figure}

A core aspect of reasoning is the ability to infer an abstract rule
from a few examples and apply it to a new instance step by step. The Abstraction and Reasoning Corpus (ARC) tests this ability
with colored-grid puzzles (Fig.~\ref{fig:teaser}, left).
Each task provides a few demonstration input--output pairs, typically
two to four, that share a hidden transformation, e.g.\ moving objects, repairing
symmetry, or recoloring by size. Given a held-out test
input, the solver must produce the output grid from scratch by
choosing its dimensions and assigning one of
ten colors to every grid cell. A prediction counts as correct only if it
matches the target grid exactly \cite{arc-agi-1}.

Because each evaluation task instantiates a rule never seen during
training, the solver must induce the transformation from its
demonstrations rather than retrieve a memorized solution
\cite{arc-agi-1}. This emphasis on generalization has made ARC a central benchmark for
abstract reasoning \cite{arc-agi-2}. Humans solve tasks from the demonstrations alone, often
constructing the answer step by step \cite{h-arc}, whereas leading ARC solvers
typically depend on large models, explicit program search, or
per-task test-time adaptation
\cite{greenblatt2024arcagi,pourcel2025self,barc,nvarc}.

Compact recurrent models tackle ARC tasks by trading scale for
iterative computation. HRM and TRM \cite{wang2025hierarchicalreasoningmodel,
jolicoeurmartineau2025morerecursivereasoningtiny} use
recurrent computation over tokenized grid sequences, whereas LoopViT \cite{shu2026loopvitscalingvisualarc}
repeatedly applies a shared visual core to spatial grid
representations, producing a decodable grid prediction at every
iteration. Yet all these models anchor supervision to the final answer,
applied once in LoopViT or repeated at every iteration in HRM and
TRM, so no signal specifies what intermediate predictions should look
like.

In principle, explicit intermediate targets would decompose a complex
transformation into simpler steps that are easier for the model to
learn. Providing
such targets requires intermediate grids from input to output, but
existing sources are insufficient: human solving traces are scarce \cite{arctraj,h-arc},
while language rationales are not directly usable as grid-state
targets \cite{arc-tgi}. Programmatic task implementations
\cite{rearc,arc-gen} offer a scalable alternative because they
implicitly encode how the output is built. However, running these programs directly does not yield clean
intermediate steps: programs may perform several changes at once or skip meaningful
intermediate states entirely. We therefore decompose and verify each program so that
execution yields
\emph{semantically monotonic transformation chains}---sequences of
intermediate grids in which each milestone applies one meaningful
action toward the answer without backtracking. These chains define what each iteration should produce.

We introduce \emph{TraceViT} (Fig.~\ref{fig:teaser}), which
trains a looped reasoner with these chains as intermediate targets.
In a looped reasoner such as LoopViT, the hidden state serves both as memory of the input and as the
computation workspace. We argue that supervising intermediate states
risks disrupting the stored input information, because the same
representation must simultaneously preserve it and undergo
transformation. To mitigate this, we
externalize the memory role through a \emph{task reference} that encodes all
demonstrations into a compact summary and re-supplies it at every
iteration, and an \emph{object workspace} that decomposes the current
scene into object-centric components and carries them across
iterations. A
separate challenge is that the number of chain steps and loop
iterations need not match. To address this, \emph{soft trace
alignment} enforces only the ordering of the chain while letting the
model decide how many iterations to spend on each step.

TraceViT achieves 67.8\% pass@2 on ARC-AGI-1 and 24.3\% on
ARC-AGI-2 (Table~\ref{tab:main}), demonstrating strong performance
across both benchmarks. These results provide empirical support for
grounded trace supervision as an effective approach to compact visual
reasoning.

In sum, we contribute:
\begin{itemize}
\item \textbf{An ARC dataset with verified trace annotations:} we
  rewrite and validate programmatic task implementations to produce
  semantically monotonic transformation chains as training targets.
  Once instrumented, trace annotations are generated alongside sampled
  instances, avoiding per-instance labeling and supporting scalable
  resampling.
\item \textbf{Grounded trace supervision:} an architecture and
  training objective that grounds every iteration with a \emph{task
  reference} and \emph{object workspace}, then uses \emph{soft trace
  alignment} to supervise intermediate predictions while letting the
  model freely allocate iterations to milestones.
\item \textbf{Systematic evaluation:} experiments demonstrate strong
  performance on both ARC-AGI-1 and ARC-AGI-2.
  A factorial ablation identifies the interaction between trace
  supervision and grounding, while targeted variants evaluate key
  alignment choices.
\end{itemize}

\section{Related Work}

\textbf{ARC-Related Datasets.}
ARC-AGI-1 \cite{arc-agi-1} and ARC-AGI-2 \cite{arc-agi-2} serve as the standard benchmarks, with each task presented as a set of input--output grid pairs. Beyond these final pairs, several datasets capture intermediate human solving steps---from action traces to low-level interaction histories \cite{h-arc,arctraj,arc-interactive-history-dataset}. While these human-collected datasets offer rich process-level supervision, they are difficult to scale. Programmatic methods instead synthesize ARC-like data automatically. RE-ARC \cite{rearc}, BARC \cite{barc}, NVARC \cite{nvarc}, and ARC-GEN \cite{arc-gen} generate new input--output pairs through procedural sampling or LLM-driven program remixing, greatly expanding available training data. ARC-TGI \cite{arc-tgi} supplements a curated subset with natural-language reasoning chains. Yet no existing resource provides intermediate grid states at the scale needed for training: human traces capture them but are scarce, while programmatic generators emit only final input--output pairs.

\textbf{Approaches to ARC.}
\emph{Program induction and transduction.} ARC solvers broadly fall into two paradigms: \emph{induction} searches for a program that maps inputs to outputs, either through symbolic DSL search \cite{dsl,arc-dsl,alford2021neural,xu2023graphs,lei2024generalized,ouellette2024efficientneurallyguidedprograminduction} or LLM-guided code generation \cite{greenblatt2024arcagi,singhal2025conceptsearch,pourcel2025self}, while \emph{transduction} directly predicts test outputs without an explicit program, typically by fine-tuning LLMs on text-serialized grids with test-time training (TTT) \cite{barc,cole2025dontthrowbabybathwater,akyurek2025surprising,franzen2025product,nvarc}. Both rely on expensive DSL search or large LLMs rather than learning a compact network that computes the transformation directly.

\emph{Recurrence and iterative refinement.} Pursuing exactly such compact models, HRM \cite{wang2025hierarchicalreasoningmodel} couples two recurrent modules at different frequencies, while TRM \cite{jolicoeurmartineau2025morerecursivereasoningtiny} reduces this to a single recursively applied network. Follow-up work adapts or varies the recursive core \cite{royeazar2026tinyrecursivemodelsarcagi1,mcgovern2025testtimeadaptationtinyrecursive,wang2026tinyrecursivereasoningmamba2}. Beyond latent-state refinement, ARChitects \cite{franzen2025architects} realizes the same iterative principle through token-level masked diffusion over serialized grids. Across these models the refinement loop is supervised only by the final answer, leaving intermediate iterations without an explicit semantic target.

\emph{Vision-native ARC.} ViTARC \cite{li2025tackling} and VARC \cite{hu2025arcvisionproblem} show that a Vision Transformer (ViT) applied directly to the grid can match text-based LLM solvers. Other recent work explores reasoning itself as a visual modality \cite{liu2026reasoningmodality,zhang2025thinkvisuallyreasontextually}. LoopViT \cite{shu2026loopvitscalingvisualarc} unifies recurrent refinement with the vision-native setting and is the architecture our method builds on. Concurrent with our work, Loop-OWM \cite{gao2026slotstransitionsloopslearning} recasts the loop as an object-centric world model in which color-prototype slots decompose the grid and a demonstration-conditioned transition model rolls the state forward. As with the looped reasoners above, its supervision anchors only at the final state, without explicit semantic targets for intermediate iterations.

\section{Method}

\begin{figure*}[t!]
\centering
\includegraphics[width=\textwidth]{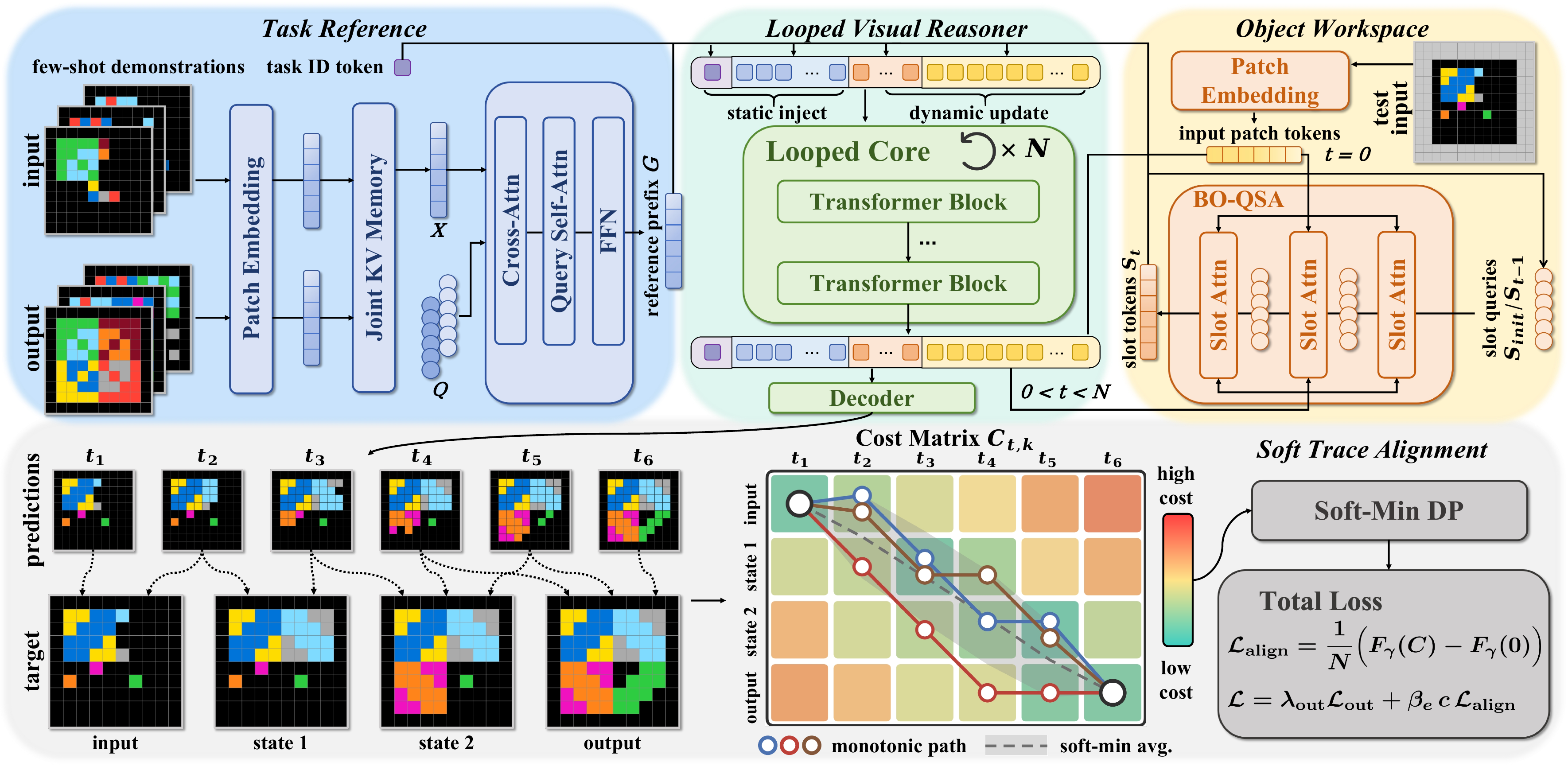}
\caption{Overview of the proposed method.
\textbf{Top:} the few-shot demonstrations are encoded into a compact
task-reference prefix $G$ (left) via cross-attention from learned
queries $Q$ to the joint demonstration context $X$; a looped visual
reasoner (center) iterates a shared core $F$ for $N$ steps. Iteration
$t$ receives the static reference $G$ and the dynamic slot bank
$S_{t-1}$; the object workspace (right) extracts $S_t$ from the
updated state for the next iteration when $t<N$.
\textbf{Bottom:} soft trace alignment computes a cost matrix
$C_{t,k}$ between the loop's decoded predictions and the
transformation chain targets. A soft-min DP marginalizes over
monotonic paths, supervising the order but not the pacing of
intermediate states; both the pairwise costs and final-state loss are
change-weighted. The indicator $c\in\{0,1\}$ marks whether an
instance is traced.}
\label{fig:method}
\end{figure*}

A looped visual reasoner processes the input over multiple iterations
before producing the final output. Under final-state supervision the
objective specifies only the endpoint but leaves the trajectory,
i.e.\ the sequence of intermediate states from input to output,
unconstrained. We therefore supervise the trajectory,
aligning each intermediate state with corresponding intermediate
targets. We propose three components to realize this principle
(Figure~\ref{fig:method}): \emph{transformation chains} define what
each iteration should produce; a \emph{task reference} and
\emph{object workspace} externalize the memory role so that
intermediate states can be freely supervised; and \emph{soft trace
alignment} enforces the ordering of the chain while letting the model
decide how to allocate its iterations.

\subsection{Preliminaries: Task Formulation and Backbone}
An ARC task contains $m$ demonstration pairs
$\{(x^{d}_{j},y^{d}_{j})\}_{j=1}^{m}$ and one or more test inputs.
Each grid has at most $30{\times}30$ cells drawn from ten colors.
Because the test inputs are answered independently, we write one
input--target pair as $(x,y)$ without loss of generality.

We adopt LoopViT \cite{shu2026loopvitscalingvisualarc}, a looped visual
reasoner, as our backbone and extend it with grounded trace
supervision. Its encoder
$E$ maps the test input and a learned task token $\tau$ to an initial
state $h_{0}$. A shared core $F$ then iterates $N$ times, producing
a sequence of hidden states $h_{t}=F(h_{t-1}+e_t)$, where $e_t$ is a
learned step embedding. A decoder maps each state to a color distribution over every grid
cell, $p_t=D(h_t)$. The conventional objective supervises only the last
prediction,
$\mathcal{L}_{\mathrm{end}}=\mathrm{CE}(p_N,y)$.
Thus, although the decoder can produce a prediction at every
iteration, only the final one receives a training signal. To
supervise the intermediate predictions as well, we first need targets
that define what each iteration should produce.

In the base model, the loop is conditioned only on the test input and
a task token supplied at initialization. Test-time training can encode
the demonstrations indirectly through weight updates, but they remain
absent from individual iterations as explicit inputs.

\subsection{Constructing Transformation Chains}
\label{sec:chains}
Trace supervision requires intermediate grids at training scale, but
existing resources do not directly provide suitable targets. Human
solving traces are scarce \cite{arctraj,h-arc}, and language
rationales describe the rule verbally without producing grid-state
targets that can supervise intermediate iterations \cite{arc-tgi}.

Programmatic task implementations offer a scalable alternative
because they implicitly encode how the output is built. RE-ARC
\cite{rearc} supplies generators and deterministic verifiers for all
400 ARC-AGI-1 training tasks, while ARC-GEN \cite{arc-gen} provides
generators for all ARC-AGI-1 tasks and 500 ARC-AGI-2 tasks. However,
running these programs directly does not yield clean intermediate
steps: programs may perform several changes at once or skip
meaningful intermediate states entirely. We therefore use GPT-5.5
with \texttt{xhigh} reasoning effort to decompose these
implementations into single-action steps.
For a traced instance, executing the decomposed program yields
\begin{equation}
x = T_{0} \;\rightarrow\; T_{1} \;\rightarrow\; \cdots
\;\rightarrow\; T_{K} = y,
\label{eq:chain}
\end{equation}
where $K=|\mathrm{steps}|\ge1$ and the final recorded stage
is the output. A chain is \emph{semantically monotonic} when each step performs
exactly one action toward the answer, without backtracking or
exploration. Here monotonic refers to the progression of actions, not to
pixel-level similarity to $y$. A single indivisible action yields
$K{=}1$. Instances for which no valid chain can be produced remain
untraced.

We validate each rewrite in two stages. Automated checks verify that the decomposed program reproduces the
original input--output pairs exactly and that every intermediate grid
is valid. We then manually review each rewritten program by visualizing
sampled chains to ensure every step corresponds to a single
meaningful action, and iterate with GPT-5.5 until all checks pass.

Correctness alone is insufficient if the rewritten generators collapse
the diversity of the training distribution. On the ARC-AGI-1 tasks
shared by ARC-GEN and RE-ARC, for example, the median number of
distinct output shapes per task is 16 for ARC-GEN and 216 for RE-ARC.
The same problem affects ARC-AGI-2. To increase diversity, we
broaden size and object-count ranges in the ARC-GEN generators and
apply a semantics-preserving color permutation to each chain. Because colors can themselves encode a rule, we manually adjudicate, for each task, whether recoloring preserves
its semantics. Examples where
recoloring changes the rule are shown in Appendix~\ref{app:recolor-examples}.

For ARC-AGI-1, we annotate RE-ARC's 400{,}000 pairs across all 400
training tasks with transformation chains, leaving inputs and
outputs unchanged; 270{,}854 of them (67.7\%) are traced.
For ARC-AGI-2, we curate a corpus of 877{,}962 instances across 891 tasks,
combining 500 ARC-GEN tasks with 391 tasks also covered by RE-ARC.
Full construction and verification details are in Appendix~\ref{app:corpus}.

\subsection{Grounding the Loop: Task Reference and Object Workspace}
Trace supervision requires intermediate states to evolve toward
chain targets, but in a looped architecture the same hidden
state also serves as the model's memory of the task. Supervising it
toward specific targets therefore risks disrupting stored input
information. This concern is further motivated by the base design,
where the task rule is captured only indirectly through test-time
weight updates
\cite{hu2025arcvisionproblem,shu2026loopvitscalingvisualarc} and
object structure is encoded only implicitly in the patch sequence. We
mitigate this by externalizing the memory role into two dedicated
structures: a static \emph{task reference} encoded from the
demonstrations and a dynamic \emph{object workspace} extracted from
the evolving recurrent state.

\paragraph{Task reference.}
We process every demonstration pair
$\{(x^{d}_{j},y^{d}_{j})\}$ through the same encoder $E$ used for the
test grid. We concatenate their non-padding patch features into a joint context
$X$, tagging each patch with three additive embeddings: a role embedding
indicating whether it comes from an input or output grid, a
demonstration-index embedding identifying which pair $j$ it belongs to,
and a two-dimensional sinusoidal positional embedding.
This joint representation allows cross-attention to compare features
across all pairs simultaneously, rather than encoding each pair in
isolation.

We encode the joint context $X$ into 128 reference tokens using
learned queries $Q=[Q_{\mathrm{a}};Q_{\mathrm{f}}]$. The 64 queries in
$Q_{\mathrm{a}}$ are initialized from an $8{\times}8$ grid of
two-dimensional positional anchors, encouraging spatial coverage; the
remaining 64 queries are unconstrained and can capture nonlocal or
task-specific relations. The queries are refined over two rounds, each applying cross-attention
into $X$, self-attention, and a feed-forward network. The resulting tokens form the reference $G$, enabling each patch token
to selectively attend to relevant demonstration evidence during the
loop. $G$ is computed once and re-injected before every loop iteration.

\paragraph{Object workspace.}
Many ARC rules select, count, move, or recolor objects
\cite{xu2024llmarc,ferre2024objectcentric}, whereas patch
tokens encode such groupings only implicitly. Slot Abstractors \cite{mondal2024slotabstractors} show that
decomposing images into object-centric slots benefits abstract visual
reasoning. Following this principle, we use Slot Attention
\cite{slot-attention} to obtain an explicit object-centric summary of
an ARC grid. By competing for patch
features, the slots iteratively decompose the scene without assuming a
fixed object segmentation.

Let $\phi$ denote this slot update, $S_{\mathrm{init}}$ its learned
initial queries, and $W_s$ a linear projection to the backbone dimension. The initial workspace is extracted before the loop, with each
subsequent workspace initialized from its predecessor:
\begin{equation}
\begin{array}{rcl}
S_{0}&=&\phi(h_{0}^{\mathrm{patch}};S_{\mathrm{init}}),\\
h_{t}&=&F\bigl(h_{t-1}+e_{t}+P_GG+P_SW_sS_{t-1}\bigr),\\
S_{t}&=&\phi(h_{t}^{\mathrm{patch}};S_{t-1}),\quad t<N.
\end{array}
\label{eq:workspace}
\end{equation}
Here $e_t$ is the step embedding of the base loop, and $P_G$ and $P_S$
write the two grounding sources into their reserved prefix
positions. We train $\phi$ with the straight-through BO-QSA update
\cite{bo-qsa}. Initializing each update from $S_{t-1}$ encourages consistent
slot assignments across iterations, though slots need not maintain a
fixed object identity.

The two grounding structures have complementary roles: $G$ supplies a
fixed description of the rule, while $S_t$ summarizes the evolving
objects. Together they make the task rule and evolving scene available
at every iteration, allowing each chain state to be interpreted in
context. The remaining question is how to align chain steps with loop
iterations.

\subsection{Supervising the Loop: Soft Trace Alignment}
Given a transformation chain, the most direct supervision strategy
assigns milestones to iterations at fixed intervals. However, some steps need more computation than others, so a fixed
assignment may be suboptimal. Instead of fixing the assignment, we require only that iterations
follow the chain in order and that the final iteration matches the
output.

For a traced instance, let $p_t=D(h_t)$ be the distribution decoded
at iteration $t$ and $\Omega_k$ the valid cells of chain state $T_k$
in Eq.~\ref{eq:chain}. We construct a pairwise cost by averaging cross-entropy over the
valid cells of each milestone:
\begin{equation}
C_{t,k}
=-\frac{1}{|\Omega_k|}
\sum_{i\in\Omega_k}\log p_t\bigl(i,T_k(i)\bigr).
\label{eq:uniform-cost}
\end{equation}
Collecting these scores over all iterations and milestones gives the
pairwise cost matrix $C$.

An alignment path is a sequence
$\pi=(\pi_0,\ldots,\pi_N)$ with $\pi_0=0$,
$0\le\pi_{t-1}\le\pi_t\le K$, and $\pi_N=K$. The path can stay at the same step but never go backward. We focus on the $K\le N$ case, where the loop has
enough iterations to visit every milestone in order and jumps larger
than one are disallowed; the $K>N$ regime, which requires skip
penalties, is detailed in Appendix~\ref{app:skips}.

Selecting only the cheapest path would lock in a single assignment
too early in training. For path $\pi$, define
\begin{equation}
A(\pi;C)=\sum_{t=1}^{N}C_{t,\pi_t}.
\label{eq:pathcost}
\end{equation}
We marginalize all admissible paths $\Pi_{N,K}$ with a
temperature-$\gamma$ free energy,
\begin{equation}
F_{\gamma}(C)=-\gamma\log\!\!\sum_{\pi\in\Pi_{N,K}}\!\!
\exp\!\left(-A(\pi;C)/\gamma\right).
\label{eq:softdp}
\end{equation}
A smaller $\gamma$ sharpens the objective toward the single best
(Viterbi) path, while a larger $\gamma$ averages over more
assignments. This
construction combines the latent monotone alignment used by CTC
\cite{graves2006ctc} with the differentiable soft-min dynamic
programming of Soft-DTW \cite{cuturi2017softdtw}, specialized here to
ordered grid states and a fixed final state.

Writing
$\mathrm{softmin}_{\gamma}\{s_j\}=-\gamma\log\sum_j e^{-s_j/\gamma}$,
the free energy is computed exactly by
\begin{equation}
V_t(k)=C_{t,k}+
\mathrm{softmin}_{\gamma}\bigl\{V_{t-1}(k{-}1),\,V_{t-1}(k)\bigr\},
\label{eq:dp}
\end{equation}
with $V_0(0)=0$, $V_0(k>0)=+\infty$, $V_t(-1)\equiv+\infty$, and
$F_{\gamma}(C)=V_N(K)$. Differentiation gives the posterior occupancy
$q_{t,k}=\partial F_{\gamma}/\partial C_{t,k}
=\Pr(\pi_t=k\mid C)$, so each decoded state receives a
posterior-weighted mixture of milestone losses rather than a fixed
target assignment.

Because the number and cost of admissible paths vary with trajectory
length, we subtract the cost under a zero matrix (with identical
transition rules) and normalize by $N$:
\begin{equation}
\mathcal{L}_{\mathrm{align}}
=\frac{1}{N}\Bigl(F_{\gamma}(C)-F_{\gamma}(\mathbf{0})\Bigr).
\label{eq:align}
\end{equation}
\begin{samepage}
Let $\mathcal{L}_{\mathrm{out}}$ denote the mean cross-entropy
between the final decoded distribution $p_N$ and the ground-truth
output $y$, averaged over valid cells. With $c\in\{0,1\}$ marking
whether an instance is traced, the training objective is
\begin{equation}
\mathcal{L}
=\lambda_{\mathrm{out}}\mathcal{L}_{\mathrm{out}}
+\beta_e\,c\,\mathcal{L}_{\mathrm{align}},
\label{eq:total}
\end{equation}
where $\beta_e$ is a warmup schedule for the alignment weight.
For untraced examples ($c{=}0$), only the final-state term applies.
\end{samepage}

The cost $C_{t,k}$ weighs every valid cell equally, but the steps it
supervises are far from uniform: the median step in our corpus changes
only 7.8\% of its cells. The uniform average is therefore dominated by
the unchanged majority, while the few cells that do change carry
nearly all of the step's semantic content. We therefore make every
per-cell cost \emph{change-weighted}: matching against milestone
$T_{k}$, cell $i$ receives weight
\begin{equation}
w_k(i)=1+\alpha\,[\,T_k(i)\ne T_{k-1}(i)\,],
\qquad w_0(i)=1,
\label{eq:weight}
\end{equation}
where $\alpha\ge0$, and replace the uniform cost with
\begin{equation}
C_{t,k}=
-\frac{\sum_{i\in\Omega_k}w_k(i)\log p_t\bigl(i,T_k(i)\bigr)}
{\sum_{i\in\Omega_k}w_k(i)}.
\label{eq:cost}
\end{equation}
The final-state term is weighted the same way:
$\mathcal{L}_{\mathrm{out}}$ averages its per-cell cross-entropy under
$w(i)=1+\alpha\,[\,y(i)\ne x(i)\,]$, which takes the test input as the
state preceding the output.

The objective has a natural starting point: early in training, all
admissible pacings contribute roughly equally, and because monotonic
paths concentrate near the diagonal, the initial supervision resembles
a smoothed uniform schedule before specializing per instance.

\section{Experiments}

Our experiments address four questions, with all controlled analyses
conducted on ARC-AGI-1. First, we compare pass@2 accuracy against
published ARC solvers on ARC-AGI-1 and ARC-AGI-2. Second, a factorial
ablation isolates the interaction between trace supervision and
grounding, while targeted variants evaluate change weighting, soft
trace alignment, and the supervision schedule. Third, we visualize how
intermediate predictions and object-workspace assignments evolve across
iterations. Finally, we decompose the oracle--pass@2 gap into candidate
coverage and selection errors to identify the dominant remaining
test-time bottleneck.

\subsection{Experimental Setup}
\textbf{Data and benchmarks.}
We evaluate on the official public benchmarks ARC-AGI-1
\cite{arc-agi-1} and ARC-AGI-2 \cite{arc-agi-2}. For ARC-AGI-1, we
retain the same official tasks and RE-ARC input--output pairs used by
prior vision-native solvers
\cite{hu2025arcvisionproblem,shu2026loopvitscalingvisualarc}. The
transformation chains are the only additional supervision. The source
contains 400{,}000 RE-ARC pairs, of which 129 are removed by the
standard $30{\times}30$ size filter. We then add 1{,}718 official
examples, yielding 401{,}589 training records. For ARC-AGI-2, the
programmatic resource contributes 877{,}962 records over 891 of the
1{,}000 training tasks. We supplement it with 4{,}308 official
examples, including examples from the 109 tasks without a
programmatic source, yielding an assembled corpus of 882{,}270
records.

\textbf{Model.}
We train two model sizes, both using an eight-block shared core
applied for $N{=}6$ loop iterations: TraceViT-Medium (width 384, 11M
parameters) and TraceViT-Large (width 512, 18M parameters).
Architectural details of the backbone, the task-reference encoder,
and the object workspace are provided in Appendix~\ref{app:architecture}.

\textbf{Training and evaluation.}
We train a separate model for each of the two benchmarks using the Adam optimizer \cite{adam}
with a learning rate of $3{\times}10^{-4}$ for 100 epochs. Because
ARC-AGI-2 is harder and its training tasks partially overlap
ARC-AGI-1's, we warm-start its model from the trained ARC-AGI-1
weights. Our evaluation follows VARC
\cite{hu2025arcvisionproblem}: each evaluation task receives 100
epochs of test-time training, with demonstration pairs augmented by
flips, rotations, and color permutations. At inference,
predictions from 510 augmented views of the test input are
de-augmented and aggregated by exact-match majority voting
\cite{akyurek2025surprising}, and the two most-voted grids form the
submissions scored by the official \emph{pass@2} metric.
Remaining hyperparameters and training details are in Appendix~\ref{app:config}.
All experiments run on a single node with 8 NVIDIA A100 80GB GPUs.

\begin{table}[t]
\centering
\setlength{\tabcolsep}{4pt}
\begin{tabular}{@{}lrcc@{}}
\toprule
System & Params & ARC-1\,$\uparrow$ & ARC-2\,$\uparrow$ \\
\midrule
\multicolumn{4}{@{}l}{\cellcolor[gray]{0.92}\itshape large language models} \\
{\deemph GLM-5.2} & {\deemph 744B} & {\deemph 77.0} & {\deemph 22.8} \\
{\deemph GPT-5.5 (xHigh)} & {\deemph n/a} & {\deemph 95.0} & {\deemph 85.0} \\
{\deemph GPT-5.4 Mini (xHigh)} & {\deemph n/a} & {\deemph 63.7} & {\deemph 18.9} \\
{\deemph Gemini 3.5 Flash (High)} & {\deemph n/a} & {\deemph 84.7} & {\deemph 72.1} \\
{\deemph Grok 4 (Thinking)} & {\deemph 1.7T} & {\deemph 66.7} & {\deemph 16.0} \\
\addlinespace
\multicolumn{4}{@{}l}{\cellcolor[gray]{0.92}\itshape recurrent models} \\
HRM             & 27M & 40.3 & 5.0 \\
TRM             & 7M & 44.6 & 7.8 \\
\addlinespace
\multicolumn{4}{@{}l}{\cellcolor[gray]{0.92}\itshape vision models} \\
VARC            & 18M & 54.5 & 8.3 \\
VARC (ensemble) & 73M & 60.4 & 11.1 \\
LoopViT (Medium)        & 11M & 63.8 & 11.5 \\
LoopViT (Large)        & 18M & 65.8 & 14.2 \\
\midrule
\textbf{TraceViT (Medium)} & 11M & 65.5 & 13.3 \\
\textbf{TraceViT (Large)}  & 18M & \textbf{67.8} & \textbf{24.3} \\
\bottomrule
\end{tabular}
\caption{Comparison on ARC-AGI-1 and ARC-AGI-2 evaluation sets. LLM results are quoted from the ARC-AGI leaderboard
\cite{arcprize2025arcagi_benchmarking}; compact-solver results are
quoted from their respective publications.}
\label{tab:main}
\end{table}

\subsection{Comparison with Prior ARC Solvers}
Table~\ref{tab:main} shows that both TraceViT variants outperform
prior compact solvers at matched parameter counts on ARC-AGI-1 and
ARC-AGI-2. ARC-AGI-2 is substantially harder: every compact solver,
ours included, sees a sharp accuracy drop. The scale of the drop is
consistent with the benchmark's design: its tasks compose multiple
interacting rules across multiple steps and define symbol meanings
within each task \cite{arc-agi-2}. A single $N{=}6$-iteration pass
must realize all of these jointly, and exact-match scoring credits
no partial composition. Compact visual reasoning on ARC-AGI-2
therefore remains an open problem.

\subsection{Ablation Studies}
\label{sec:attribution}
\begin{table}[t]
\centering
\setlength{\tabcolsep}{0pt}
\begin{tabular*}{\columnwidth}{@{\extracolsep{\fill}}lcc@{}}
\toprule
 & \multicolumn{2}{c@{}}{Grounding structures} \\
\cmidrule(l){2-3}
Trace supervision & absent & present \\
\midrule
none (final state only) & 62.3 & 63.9 \\
chain milestones & 61.6 & \textbf{65.5} \\
\addlinespace
\multicolumn{3}{@{}l}{\cellcolor[gray]{0.92}\itshape variants of the full model} \\
\multicolumn{2}{@{}l}{uniform alignment cost ($\alpha=0$)} & 64.3 \\
\multicolumn{2}{@{}l}{fixed-interval alignment} & 64.4 \\
\multicolumn{2}{@{}l}{late-stage trace relaxation} & 63.6 \\
\bottomrule
\end{tabular*}
\caption{Ablation results (pass@2 \%, TraceViT-Medium, ARC-AGI-1).
The $2{\times}2$ block varies two factors: whether milestone
supervision ($\mathcal{L}_{\mathrm{align}}$) is added to the
final-state loss ($\mathcal{L}_{\mathrm{out}}$), and whether
task-reference and object-workspace grounding is enabled. The three
additional rows each change one design choice: uniform
cost sets $\alpha=0$ in Eq.~\ref{eq:weight}, fixed-interval
alignment replaces soft alignment, and late-stage trace relaxation
sets $\beta_e=0$ after epoch 40.}
\label{tab:ablation}
\end{table}
Table~\ref{tab:ablation} disentangles the contributions of trace
supervision and grounding. The $2{\times}2$ block crosses two
factors: whether training uses only the final-state term
$\mathcal{L}_{\mathrm{out}}$ or adds the milestone-alignment term
$\mathcal{L}_{\mathrm{align}}$, and whether the task reference and
object workspace are jointly enabled. Three additional variants each
modify one aspect of the full model: setting $\alpha=0$ in the
milestone cost, replacing soft alignment with a fixed-interval
schedule, or setting $\beta_e=0$ after epoch 40. All seven
configurations use TraceViT-Medium and the same ARC-AGI-1 training
and evaluation protocol.

\paragraph{Trace--grounding interaction.}
The $2{\times}2$ block reveals a clear asymmetry: trace supervision
degrades accuracy without grounding but becomes beneficial once
grounding is present. Starting from the ungrounded, final-state-only
baseline (62.3\%), adding milestone supervision alone lowers pass@2
to 61.6\%. Adding the task-reference encoder and Slot Attention workspace
alone raises it to 63.9\%, consistent with the expected benefit of
re-supplying the task rule and scene at every iteration. Combining
both yields 65.5\%, a substantial gain over either single-factor
variant. We hypothesize that this interaction arises because the recurrent
state in LoopViT serves both as memory of the input and as the
computation workspace. Under milestone supervision, intermediate states are pushed toward
partial outputs, which may disrupt the input information that
subsequent iterations depend on to complete the transformation. The task reference and object
workspace externalize this memory, making the rule and scene available
at every iteration so that the recurrent state need not carry them.
Once memory is decoupled from prediction, trace supervision guides
the transformation without starving later iterations of input
information.

\paragraph{Data control.}
The grounding-only and full-model configurations use exactly the
same input--output pairs, differing only in whether the chain
milestones enter the trace loss. The 1.6-point gap
therefore estimates trace supervision's contribution with
training-set size and diversity held constant.

\paragraph{Change weighting.}
Setting $\alpha=0$ in Eq.~\ref{eq:weight} restores the uniform
milestone cost of Eq.~\ref{eq:uniform-cost} while retaining all other
components. Pass@2 drops from 65.5\% to 64.3\%, indicating that
without change weighting the unchanged majority of cells dominates the
alignment cost.

\paragraph{Alignment rule.}
Replacing soft alignment with a fixed-interval schedule while keeping
the milestones and grounding unchanged lowers pass@2 from 65.5\% to
64.4\%, confirming the benefit of letting the model decide how to allocate
iterations to milestones.
Soft alignment also remains applicable when $K$ exceeds the iteration
budget, whereas the fixed schedule must omit milestones.

\paragraph{Trace-supervision schedule.}
Setting $\beta_e=0$ after epoch 40 removes the milestone-level
constraint for the remainder of training. Pass@2 drops from 65.5\% to
63.6\%, indicating that trace supervision is not merely an early
optimization scaffold but remains beneficial when maintained
throughout.

\subsection{Process Visualization}
\begin{figure}[t]
\centering
\includegraphics[width=\columnwidth]{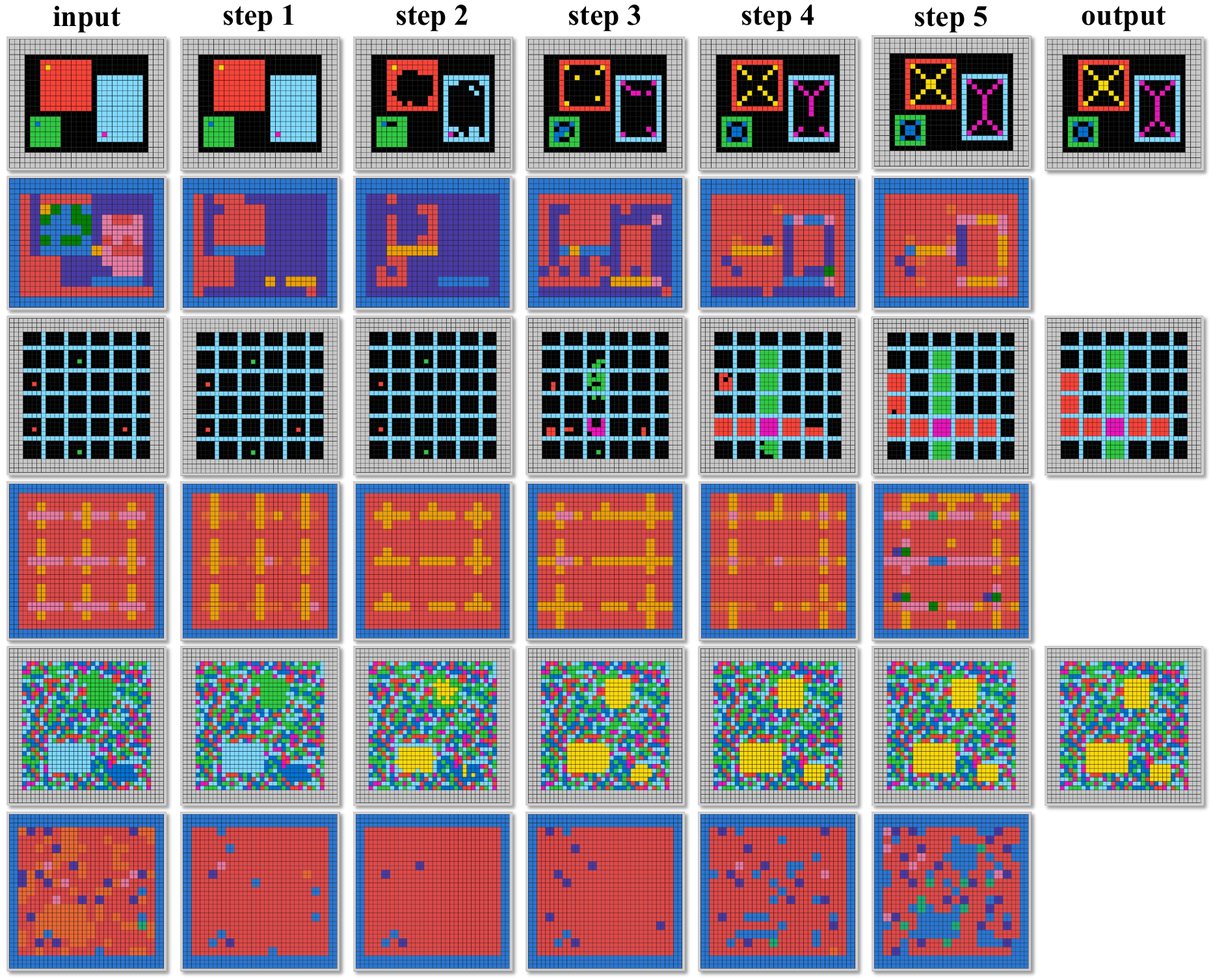}
\caption{\textbf{Per-step predictions and slot assignments} on three tasks. Prediction rows show the input and six loop
outputs. Assignment rows show the argmax over eight slots at each
step.}
\label{fig:execute}
\end{figure}

\paragraph{Evolution of intermediate predictions.}
Figure~\ref{fig:execute} shows per-step predictions and
slot assignments on three tasks. The model builds its answer incrementally across
iterations rather than producing it in one step. In the top example,
three colored boxes must each be filled with a pattern determined by a
single seed cell in their interior. The model solves this in two
visible phases: it first clears the box interiors (steps 1--2), then
draws each box's pattern (steps 3--5). The middle and bottom examples
show similar progressive construction, with each iteration refining a
different aspect of the output.

\paragraph{Evolution of workspace organization.}
Across all tasks we inspected, two slots play fixed roles: one covers
the empty canvas outside the grid and another covers the grid
background. The remaining slots carve out task-specific regions that
correspond to where the transformation happens. In the top example, the
clearing phase shows a coarse spatial bipartition across slots. When
the model switches to drawing, one slot expands to cover both box
interiors while the others narrow down to box boundaries and
pattern-row segments. This suggests that the slots shift from coarse spatial roles in the
clearing phase to fine-grained roles once drawing begins. Similar phase-aligned slot transitions appear in the other two
examples; see Appendix~\ref{app:visualization} for additional visualizations.

\subsection{Error Analysis}

\paragraph{Coverage and selection errors.}
For each ARC-AGI-1 evaluation task, we merge identical predictions
across augmented views and rank the resulting candidates by the vote
count used for submission. Let $r$ denote the rank of the correct
grid, with $r=\infty$ when it is absent. Pass@2 succeeds when
$r\leq2$, whereas oracle accuracy succeeds whenever $r<\infty$.
Thus, $2<r<\infty$ represents a selection gap, while $r=\infty$
represents a coverage error.

\begin{table}[t]
\centering
\begin{tabular*}{\columnwidth}{@{\extracolsep{\fill}}lccccc@{}}
\toprule
 & pass@2 & \multicolumn{2}{c}{oracle only} & & total \\
\cmidrule(lr){2-2}\cmidrule(lr){3-4}\cmidrule(l){6-6}
Rank $r$ & 1--2 & 3--10 & $>$10 & absent & Oracle acc. \\
\midrule
Tasks (\%) & 67.8 & 5.0 & 3.5 & 23.8 & 76.3 \\
\bottomrule
\end{tabular*}
\caption{Rank of the correct grid in TraceViT-Large's vote-ranked
candidate pool on 400 ARC-AGI-1 tasks. Finite ranks above 2 form the
selection gap, whereas ``absent'' denotes a coverage error. Oracle
accuracy sums all finite-rank bins; percentages are rounded.}
\label{tab:rank}
\end{table}

\paragraph{Coverage dominates.}
As shown in Table~\ref{tab:rank}, 67.8\% of tasks place the correct
grid in the top two, while another 8.5\% generate it at lower ranks,
raising oracle accuracy to 76.3\%. The correct grid is absent for the
remaining 23.8\%, so coverage accounts for nearly three times as much
residual error as selection, and reranking can recover at most the
8.5-point selection gap. The asymmetry is structural: each view is
decoded deterministically, so voting can aggregate correct predictions
but never generate new ones. When the correct grid is absent from all
510 views, this reflects a competence gap, not a sampling one. Where
the solver does succeed, voting already places the correct grid in
the top two for most oracle-accessible tasks (67.8 of 76.3 points).
Improving coverage must therefore come from the per-view solver
itself, consistent with the single-view bottleneck reported by VARC
\cite{hu2025arcvisionproblem}.

\section{Conclusion}
Looped visual reasoners produce intermediate predictions at every
iteration, yet endpoint-only training imposes no structure on them.
We introduced TraceViT to supervise these predictions with explicit
intermediate targets: verified, semantically monotonic transformation
chains define the milestones; a task reference and object workspace
ground every iteration; and soft trace alignment enforces milestone
order without fixing the model's pace. TraceViT achieves 67.8\%
pass@2 on ARC-AGI-1 and 24.3\% on ARC-AGI-2. Controlled ablations on
ARC-AGI-1 reveal a clear interaction: trace supervision alone slightly
degrades accuracy, yet combining it with grounding yields a
substantial gain over either component in isolation. This confirms
that supervising intermediate states requires dedicated structures to
preserve input information throughout the loop. An error decomposition
further shows that candidate coverage, not selection, is the dominant
remaining bottleneck. Grounded trace supervision thus
offers a principled way to leverage the intermediate predictions that
looped architectures already produce. The current approach assumes
that reliable chains can be obtained from programmatic task
implementations. Extending grounded trace supervision to tasks where
intermediate states must be inferred, such as through learned
decomposition or human demonstrations, is an important open
direction. More broadly, the principle of supervising iterative
computation with structured intermediate targets may generalize to
other domains where models refine predictions over multiple steps.

\bibliography{aaai2027}

\twocolumn[%
  \begin{center}
    {\LARGE\bf Appendix\par}
    \vspace{1.2em}
  \end{center}%
]
\pdfbookmark[1]{Appendix}{appendix-banner}%
\appendix
\setcounter{secnumdepth}{2}

\section{Chain-Corpus Construction Details}\label{app:corpus}

This section expands the chain-corpus construction pipeline summarized
in the Method section of the main paper. We first describe how the
ARC-GEN generators and RE-ARC verifiers are rewritten to expose
semantically meaningful intermediate states, followed by automated
checks and human review that preserve functional fidelity and step
coherence. We then detail constrained recoloring and parameter
widening for greater diversity, before presenting the deterministic
corpus assembly and final audit.

\subsection{Source Programs}
Our corpus builds on two complementary program collections. RE-ARC
\cite{rearc} pairs each of the 400 ARC-AGI-1 training tasks with a
generator that samples fresh instances and a verifier that
deterministically derives the output from the input. The released
collection contains 1{,}000 sampled pairs per task and spans diverse
sizes, palettes, and layouts, but provides no process supervision and
does not guarantee reproduction of the official examples.

ARC-GEN \cite{arc-gen} provides generators for 900 tasks: all 400
ARC-AGI-1 training tasks and 500 ARC-AGI-2 training tasks. Under its
fidelity contract, each generator reproduces its task's official
examples cell for cell when invoked with a fixed parameter setting.

The two collections encode solution structure in different
components. ARC-GEN's generators follow ``construct the puzzle, then
the output,'' whereas RE-ARC's verifiers follow ``derive the output
from the puzzle.'' We rewrite only these components. Both ordinarily
emit the final output in one shot, without exposing intermediate
states. Importantly, we never modify RE-ARC's generators or its
400{,}000 released pairs.

\subsection{Rewriting Principles}
We use GPT-5.5 with \texttt{xhigh} reasoning effort to produce every
rewrite. For each task, the model receives the source program, a
description of the task rule, and the fixed instructions below. Later
batches additionally receive reusable instructions distilled from
recurring faults in earlier reviews.
\begin{itemize}
\item \textbf{Distribution invariant.} The random sampling logic and
public interface remain unchanged; only the construction of the output
may be reorganized.
\item \textbf{Input first, output progressively.} The input is
built by the original logic; the output starts as a blank grid (or a
copy of the input) and is constructed progressively.
\item \textbf{One transition, one nameable action.} Each successive
frame must result from a single human-nameable action (``outline every
red frame,'' ``move the $k$-th object to the bottom''). We judge the
semantic operation rather than the number of changed pixels: placing
a one-cell marker may be a valid step, whereas a no-op initial frame
is not.
\item \textbf{Per-object unrolling with capacity guards.} When the
object count is sampled, repeated phases unroll one object at a time.
Any widened range must be covered by explicit capacity guards;
otherwise, overflow can silently drop objects (see \emph{Parameter
Widening} below).
\item \textbf{Frame budget.} Typical chains span two to nine frames
including the final output. A one-frame chain is permitted only for a
genuinely atomic transformation and is flagged for review.
\item \textbf{No fabricated traces.} Tasks without a meaningful
transformation chain remain untraced rather than being assigned
artificial frames. A traced, one-frame chain records one meaningful
atomic action with $K{=}1$; an untraced sample receives only direct
endpoint supervision.
\end{itemize}

\subsection{Automated Verification Gates}
The rewrite criteria define the desired chain structure; six
automated gates then test functional fidelity and equivalence. Any
failure returns the rewrite to the model for correction.
\begin{enumerate}
\item \emph{Fidelity.} An ARC-GEN rewrite reproduces its task's
official examples cell for cell at the contract parameters. A RE-ARC
rewrite recomputes all 1{,}000 released pairs for its task; across 400
tasks, this covers 400{,}000 pairs with no mismatches or execution
errors.
\item \emph{Old--new equivalence.} Across hundreds of seeds, the
rewritten and original programs produce identical outputs and
identical error and rejection behavior. This tests that the rewrite
reorganizes the existing computation rather than changing the data
distribution.
\item \emph{Structural invariants.} Every recorded frame is a
rectangular grid of colors 0--9 with dimensions at most
$30\times 30$. For traced examples, the final frame equals the output;
untraced examples carry no process targets.
\item \emph{Frame-count monitoring.} We track the frame-count
distribution of every task. Fragmentation (abnormally many frames) or
collapse (concentration on a single frame) triggers review.
\item \emph{Degeneracy and solvability.} We flag constant outputs,
identity outputs, and input collisions. An input collision makes the
generated relation inconsistent because the same input is paired with
two different outputs. Tasks whose rules genuinely permit identity or
constant outputs receive explicit exemptions.
\item \emph{Independent re-verification.} We do not rely on the
model's self-reported test results. Instead, we re-run every gate and
fingerprint the codebase between batches to detect out-of-scope edits
and silent rollbacks.
\end{enumerate}

\subsection{Human Review and Iterative Correction}
Automated gates establish functional correctness but cannot determine
whether the intermediate states form a semantically coherent
transformation chain. We therefore render every gate-passing rewrite
as an input$\,\rightarrow\,$steps$\,\rightarrow\,$output filmstrip,
one page per task. To avoid clusters of nearly identical samples, we
sample broadly across the variant space.

Review asks three questions: should adjacent frames be merged because
they fragment a single action; should any transition be split because
it fuses several actions; and can each transition be named as a
solution action so that the full chain forms a coherent narrative? We
judge the rendered frames rather than the model's accompanying
explanations, which can disagree with the program's actual behavior.
Because no automated gate can assess step semantics, human review is
the final adjudicator.

We record task-specific comments and return them to the model with the
current program; every revised program re-enters the full verification
battery. Recurring faults become reusable instructions for later
batches. For example, per-object phases are the preferred granularity;
counting tasks clear distractors before producing the count; long
paths are segmented at crossing events; trailing one- or two-cell
marks join the final phase rather than forming separate steps; and
degenerate puzzles are excluded from sampling when the official
examples always contain the triggering elements. Most batches
converged after one correction round, with roughly a dozen fixes per
50-task batch.

\subsection{Diversity: Constrained Recoloring}
Functional correctness does not by itself guarantee distributional
diversity. Among tasks covered by both collections, the median number
of distinct output grid shapes per task is 16 for ARC-GEN and 216 for
RE-ARC; the corresponding medians for distinct palettes are 14 and
162. ARC-GEN is likewise less diverse in object counts and densities.
We therefore augment color diversity with full-trajectory bijections,
applying the same mapping to the input, every intermediate frame, and
the output while handling foreground and background colors separately.

Not every color permutation preserves task semantics. Specific colors
may encode the rule, as in a fixed color-to-shape legend, a gray axis,
or a black corridor. Remapping such colors can produce an internally
consistent trajectory that is nevertheless invalid under the original
task rule. We therefore impose task-specific recoloring constraints.

Across the 500 ARC-AGI-2 tasks, 80 disable recoloring entirely, whereas
324 freeze at least one load-bearing color: the color is neither
replaced nor used as the target of another color. Background
constraints are tracked separately. The black background is locked in
45 tasks because it carries content semantics, while 113 tasks have
non-black backgrounds that may be recolored.

Automated probes help propose these constraints but err in both
directions. Treating every color mentioned by a program as fixed is
overly restrictive, whereas testing only whether a remapped trajectory
is internally consistent misses load-bearing colors. We therefore
manually adjudicate the recoloring policy for every task. As review
progressed, we refined the background criterion: black is locked only
when it carries content semantics in the \emph{input}.

\subsection{Recoloring Failure Examples}
\label{app:recolor-examples}

\begin{figure*}[p]
\centering
\includegraphics[
  width=\textwidth,
  height=0.86\textheight,
  keepaspectratio,
  trim=50 54 50 50,
  clip
]{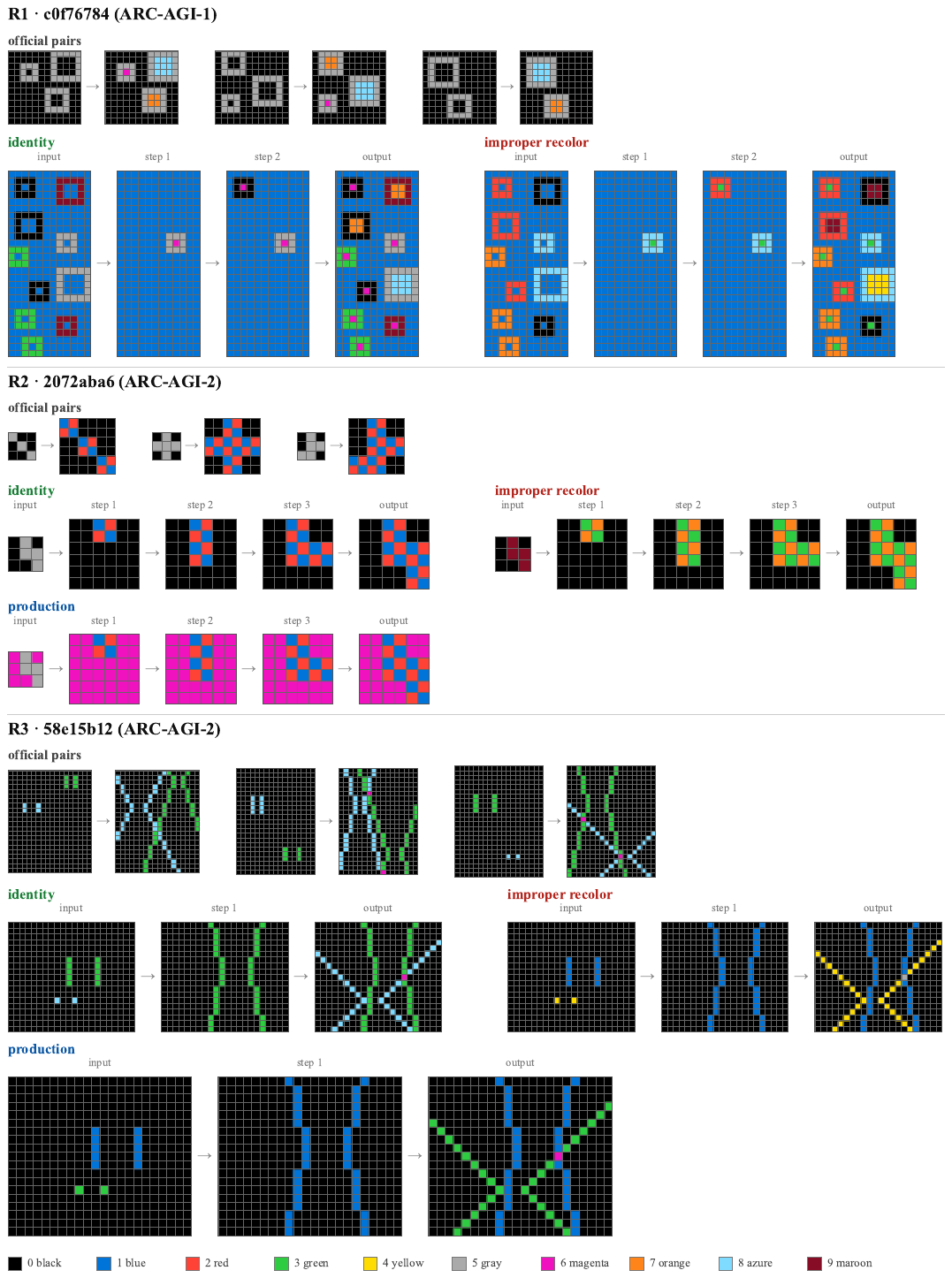}
\caption{\textbf{Failure cases for trajectory-wide recoloring.}
Official pairs establish the task semantics. Records labeled
\emph{identity} or \emph{production} preserve those semantics,
whereas each \emph{improper recolor} applies a consistent palette
mapping across the full trajectory but changes the rule encoded by
the colors.}
\label{fig:recolor-failures}
\end{figure*}

Figure~\ref{fig:recolor-failures} provides three examples.
In each case, applying one permutation consistently to the input,
intermediate states, and output preserves the appearance of a valid
trajectory but changes a color-dependent rule established by the
official examples.

\subsection{Diversity: Parameter Widening}
Beyond recoloring, we increase content diversity by widening generator
parameter spaces. We decouple height and width from shared size
parameters, broaden size ranges, expose and randomize sizes previously
hidden in default values, and parameterize object counts.

Every widening must preserve byte-level fidelity along the default
parameter path, so official examples remain unchanged and each edited
program re-enters the full verification battery. The principal risk is
silent object loss. Increasing one dimension can push derived
quantities, such as diagonal counts or sub-block counts, beyond the
original implementation's capacity. The program may then produce a
structurally valid but rule-incomplete output that no structural gate
detects; explicit capacity guards are therefore mandatory.

Recoloring policies and parameter-widening edits are both made on a
per-task basis, proceed in the same batches, and undergo the same
filmstrip review as the step rewrites.

\subsection{Corpus Assembly and Audit}
With all available programs rewritten and diversified, we synthesize
the corpus in a single deterministic pass. Each task draws from a fixed
seed stream, and every record carries complete variant metadata, making
the corpus reproducible record by record and resumable after
interruption. We deduplicate records by \((x,y)\), impose a timeout on
each attempt, and assign each task a time budget. Tasks with
heavy-tailed runtimes may therefore produce fewer records; every
shortfall is recorded in the corpus ledger.

Official examples serve only as fidelity targets in the construction
pipeline: ARC-GEN generators must reproduce them at the contract
parameters, but the examples themselves are not included in the corpus
contribution. The RE-ARC portion is pure post-processing over the
existing 400{,}000 pairs: 270{,}854 instances (67.7\%) are traced.

The resulting ARC-AGI-2 corpus contains 877{,}962 records across 891
of the 1{,}000 training tasks: 500 tasks from ARC-GEN synthesis and
391 ARC-AGI-2 tasks that overlap ARC-AGI-1 and are therefore covered
by RE-ARC. The remaining 109 tasks have no programmatic source and are
not represented in the contributed corpus. Among the 500 ARC-GEN
tasks, 499 yield traces; the remaining task is untraced because it has
no meaningful intermediate state.

Finally, we audit the complete corpus. The audit re-runs all structural
invariants and confirms that the training loader drops no records,
reconciles per-task counts with the ledger, re-checks byte-level
fidelity of RE-ARC records against their source, revalidates ARC-GEN
generators against the official JSON without adding those examples to
the corpus, verifies every recoloring constraint on the final records,
and confirms that official-example files are not ingested as records
in the contributed corpus.

\section{Alignment When Milestones Outnumber Iterations}
\label{app:skips}

When a traced chain has more milestones than the loop has iterations,
$K>N$, no non-decreasing path from chain state $0$ to chain state $K$
can visit every state in $N$ steps. Skips are therefore unavoidable.
We extend the alignment formulation in the main text to this regime by
allowing skip transitions and assigning them an explicit penalty.

We retain the main text's endpoint-anchored paths
$\pi=(\pi_0,\ldots,\pi_N)$, with $\pi_0=0$,
$0\le\pi_{t-1}\le\pi_t\le K$, and $\pi_N=K$, but now allow an
iteration to advance by more than one milestone. Let $\Pi_{N,K}$
denote this enlarged path set. Each transition incurs a penalty
proportional to the number of milestones it skips:
\begin{equation}
\Delta(\pi)=\lambda_{\mathrm{skip}}\sum_{t=1}^{N}
\max\bigl(\pi_t-\pi_{t-1}-1,\;0\bigr),
\label{eq:app-skip}
\end{equation}
where $\lambda_{\mathrm{skip}}$ is the cost per skipped milestone.
The path energy in this regime is therefore
$A(\pi;C)=\sum_{t=1}^{N}C_{t,\pi_t}+\Delta(\pi)$. The first term
measures the agreement between each iteration and its assigned
milestone, whereas the second makes every omission explicit and
costly.

Including the skip penalty in each path energy gives
\begin{equation}
F_{\gamma}(C)=-\gamma\log\!\sum_{\pi\in\Pi_{N,K}}
\exp\!\left(-A(\pi;C)/\gamma\right).
\label{eq:app-softdp}
\end{equation}
Correspondingly, the dynamic-programming recursion must consider
every predecessor $0\le j\le k$:
\begin{equation}
V_{t}(k)=C_{t,k}
+\mathop{\mathrm{softmin}_{\gamma}}_{0\le j\le k}
\bigl[V_{t-1}(j)+\lambda_{\mathrm{skip}}\max(k{-}j{-}1,0)\bigr],
\label{eq:app-dp}
\end{equation}
with $V_0(0)=0$, $V_0(k>0)=+\infty$, and
$F_{\gamma}(C)=V_N(K)$.

The main-text recursion is the $K\le N$ specialization of
Eq.~\ref{eq:app-dp}. In that regime, jumps larger than one are
disallowed, so $\pi_t-\pi_{t-1}\in\{0,1\}$, every term of
$\Delta(\pi)$ vanishes, and the predecessor set reduces to
$\{k{-}1,k\}$. The batched implementation evaluates every
$0\le j\le k$ in both regimes; when skips are disallowed,
transitions with $j<k{-}1$ receive infinite cost.

The normalization baseline must use the same transition rules and
penalties. Consequently, $F_{\gamma}(\mathbf{0})$ reflects both path
multiplicity and skip penalties rather than path multiplicity alone.
The normalized alignment loss remains zero when $C=\mathbf{0}$. We
use $\lambda_{\mathrm{skip}}=0.3$ throughout.

In this regime, order-only supervision is necessary if every
milestone is to remain eligible without prescribing omissions in
advance. A fixed schedule must preselect which milestones to omit. By
contrast, a latent monotone alignment---soft or hard---keeps every
milestone eligible as a supervision target, lets the path determine
which milestones to skip for each instance, and charges every
omission.

Allowing skips also admits the limiting case corresponding to deep
supervision. The path $\pi_0=0$ and $\pi_t=K$ for all $t\ge1$ is
valid, although its first transition incurs a skip penalty. If the
posterior occupancy concentrates on this path, every iteration is
supervised by the final target.

\section{Model Architecture Details}
\label{app:architecture}

This section details the two model sizes used in the main paper,
TraceViT-Medium (11M) and TraceViT-Large (18M). We first describe
their shared backbone and recurrent loop, then the task-reference
encoder and the object workspace.

%
%
\begin{table*}[t]
\centering
\small
\setlength{\tabcolsep}{2pt}
\begin{tabular}{@{}ll@{\enskip}ll@{\enskip}ll@{}}
\toprule
\multicolumn{2}{@{}l}{\itshape Offline training} &
\multicolumn{2}{l}{\itshape Loss} &
\multicolumn{2}{l@{}}{\itshape Test-time training (per task)} \\
\midrule
Optimizer & Adam
  & Final-state weight $\lambda_{\mathrm{out}}$ & 2.0
  & Initialization & EMA weights of the run \\
Learning rate & $3\times10^{-4}$, cosine
  & Alignment weight $\beta$ & 0.2 (Medium), 0.3 (Large)
  & Epochs & 100 \\
Warmup & 10 epochs, linear
  & Alignment warmup & 5 epochs, linear
  & Batch size & 8 \\
Batch size & 32 per GPU (256 total)
  & Softmin temperature $\gamma$ & 0.5
  & Learning rate & $3\times10^{-4}$, cosine \\
Epochs & 100
  & Skip penalty $\lambda_{\mathrm{skip}}$ & 0.3
  & Precision, clipping & bf16, 1.0 \\
Gradient clipping & 1.0
  & Change weight $\alpha$ & 3.0
  & Objective & final-state CE, $\alpha{=}0$ \\
Precision & bf16 autocast
  & &
  & Independent runs & 2 \\
Dropout & 0.1
  & &
  & Views per test input & $51\times10$ per run \\
Weight averaging & EMA, decay 0.9999
  & & & & \\
Seed & 42
  & & & & \\
\bottomrule
\end{tabular}
\caption{Training and evaluation configuration. Every value is shared
by the two model sizes and by the two benchmarks except $\beta$ and
the ARC-AGI-2 warm-start changes described in the text.}
\label{tab:app-config}
\end{table*}

\subsection{Backbone and Loop Configuration}
\label{app:backbone}

\paragraph{Canvas and tokenization.}
Every grid is rendered on a fixed $64\times64$ canvas. A grid of at
most $30\times30$ cells is upscaled by an integer factor using
nearest-neighbor interpolation and placed at an offset within the
canvas. Training and test-time inference resample the scale and offset
on every pass, whereas offline evaluation and demonstration rendering
use fixed placements.

Positions outside the grid carry a dedicated background index. Every
target canvas---the answer or an intermediate chain state---also
carries a one-cell border of a second dedicated index along the grid's
right and bottom edges. This border lets the model encode the predicted
grid shape directly in the canvas instead of using a separate shape
predictor. The resulting vocabulary contains twelve symbols: the ten
ARC colors, background, and border. The loss is evaluated only on the
grid-and-border region. Within a traced instance, the input,
intermediate states, and output use the same scale and offset, keeping
the entire trajectory pixel-aligned.

\paragraph{Core block.}
Both model sizes use the pre-norm hybrid block of LoopViT
\cite{shu2026loopvitscalingvisualarc}:
\[
\begin{array}{rcl}
x&\leftarrow&x+\mathrm{MHSA}\bigl(\mathrm{RMSNorm}(x)\bigr),\\
x&\leftarrow&x+\mathrm{ConvGLU}\bigl(\mathrm{RMSNorm}(x)\bigr).
\end{array}
\]
The ConvGLU branch uses a nominal feed-forward width of 512, giving a
gated hidden width of $\lfloor 2\cdot512/3\rfloor=341$. It projects
each token to twice this hidden width and splits the result into a gate
and a value. For image tokens, the gate is reshaped to the
spatial patch grid and processed by a $3\times3$ depthwise
convolution; prefix tokens bypass this convolution. The branch then
recombines gate $g$ and value $v$ as
$\mathrm{GELU}(g)\odot v$ before the output projection.

\paragraph{Loop.}
The core is unrolled for exactly $N{=}6$ iterations. Each iteration
performs the following operations in order:
\begin{enumerate}
\item add the step embedding $e_t$ to the whole sequence, prefix
included;
\item add the reference $G$ at its reserved positions;
\item add the projected workspace $W_sS_{t-1}$ at its reserved
positions;
\item apply the eight shared blocks;
\item decode the current prediction; every iteration is decoded during
training because the alignment cost matrix requires all six outputs;
\item re-extract the workspace from the updated patch tokens, except
after the final iteration, whose slots would never be re-injected.
\end{enumerate}
Under the fixed unroll, the prediction decoded at $t=N$ is the model's
output. Sharing block parameters across iterations yields
$6\times8=48$ block applications per forward pass while storing the
parameters of only eight blocks.

\paragraph{Decoder.}
The decoder first normalizes the patch tokens and maps each one through
Linear$\,\rightarrow\,$GELU$\,\rightarrow\,$Linear, producing
$12\cdot2^{2}=48$ logits per token. De-patchification converts these
values into a $12\times64\times64$ logit map. The decoder runs at every
iteration. Consequently, the intermediate
predictions used by the alignment loss and the final answer are
produced by an identical decoding path.

\paragraph{The two sizes.}
The two models differ only in embedding width: 384 for
TraceViT-Medium and 512 for TraceViT-Large. Both use an eight-block
core, six loop iterations, eight attention heads, and a nominal
feed-forward width of 512. Their object workspaces contain eight slots,
run three internal rounds, and use a 256-wide projection space. Their
task-reference encoders produce 128 tokens with a 128-wide, two-round
encoder over at most four demonstrations.

The per-task token table is the only parameter component that does not
transfer. It is discarded and re-initialized for test-time training,
and the parameter counts reported in the main paper exclude it.

\subsection{Task-Reference Encoder}
\label{app:reference}

The task-reference encoder converts at most four demonstration pairs
into the static reference $G$. Each input and output grid is rendered
at a canonical placement on the $64\times64$ canvas and processed by
the backbone's shared color, patch, and positional embeddings. After a
projection to the encoder width $d_{\mathrm{enc}}{=}128$, a token from
demonstration $j$, stream $s\in\{\mathrm{in},\mathrm{out}\}$, and patch
position $\ell$ is represented as
\begin{equation}
X_{j,s,\ell}=W_{\mathrm{in}}H^{s}_{j,\ell}
+\mathrm{tag}^{\mathrm{stream}}_{s}
+\mathrm{tag}^{\mathrm{demo}}_{j}
+\mathrm{pos}_{\ell},
\label{eq:app-field}
\end{equation}
where the stream and demonstration tags are learned and
$\mathrm{pos}_{\ell}$ is a fixed two-dimensional sine--cosine code.
Valid input and output tokens from all demonstrations are flattened
into the joint context $X$.

\paragraph{Reference extraction and injection.}
$Q=[Q_{\mathrm{a}};Q_{\mathrm{f}}]$ contains 128 queries. The 64
queries in $Q_{\mathrm{a}}$ are initialized with the two-dimensional
sine--cosine code of an $8\times8$ grid, while the 64 queries in
$Q_{\mathrm{f}}$ are freely learned. Two four-head rounds update the
queries by
\begin{equation}
\begin{array}{rcl}
q&\leftarrow&\mathrm{LN}\bigl(q+\mathrm{CrossAttn}(q,X)\bigr),\\
q&\leftarrow&\mathrm{LN}\bigl(q+\mathrm{SelfAttn}(q)\bigr),\\
q&\leftarrow&\mathrm{LN}\bigl(q+\mathrm{FFN}(q)\bigr),
\end{array}
\label{eq:app-xattn}
\end{equation}
where the feed-forward hidden width is $2d_{\mathrm{enc}}$ and the
activation is GELU. A final linear map lifts the refined queries to the
backbone width, producing $G$. The reference is computed once,
concatenated into its reserved prefix positions in the initial state,
and added again at those positions before every loop iteration. This
additive re-injection restores the same demonstration evidence without
discarding the prefix state accumulated by the loop.

\subsection{Object Workspace}
\label{app:slots}

The object workspace applies Slot Attention \cite{slot-attention} to
the valid patch features of the recurrent state. Let
$Z\in\mathbf{R}^{B\times P\times d}$ denote these features on the fixed
canvas and $m\in\{0,1\}^{B\times P}$ their validity mask. Each
extraction projects normalized features once into keys $K$ and values
$V$ and maintains eight slots $S$. Before the loop, the slots are
initialized from learned queries; subsequent extractions start from
the workspace produced by the preceding loop iteration.

\paragraph{Competitive slot update.}
For each internal round, queries
$Q=W_q\mathrm{LN}(S)$ produce scaled slot--patch affinities
$\ell_{b,k,n}=\langle Q_{b,k},K_{b,n}\rangle/\sqrt{d}$. A softmax over
the eight slots makes them compete for each patch, after which the
validity mask removes positions outside the grid:
\begin{equation}
\begin{array}{rcl}
\tilde a_{b,k,n}
&=&m_{b,n}\,
\frac{\exp(\ell_{b,k,n})}{\sum_j\exp(\ell_{b,j,n})},\\
u_{b,k}
&=&\displaystyle\sum_n
\frac{\tilde a_{b,k,n}}
     {1+\sum_{n'}\tilde a_{b,k,n'}}\,V_{b,n}.
\end{array}
\label{eq:slot-aggregate}
\end{equation}
The added unit in the denominator suppresses updates to slots that
receive little attention. A shared GRU combines $u$ with the previous
slot state, followed by a residual MLP.

\paragraph{Iterations and loop integration.}
Each extraction uses three update rounds. Following BO-QSA
\cite{bo-qsa}, the first two rounds do not record gradients, and
\[
\tilde S^{(2)}
=\mathrm{sg}\!\left(S^{(2)}\right)+S^{(0)}
-\mathrm{sg}\!\left(S^{(0)}\right)
\]
provides a straight-through connection to the third,
gradient-enabled round. At each loop iteration, the current slots are
projected by $W_s$ and added at the reserved workspace positions; the
next workspace is then extracted from the updated patch states,
initialized from the current slots. No extraction follows the final
iteration.

\section{Training and Evaluation Configuration}
\label{app:config}

This appendix records the training configuration and the test-time
protocol.
Table~\ref{tab:app-config} collects the values; the paragraphs below
give the parts that a value alone does not specify. Both model sizes
share every setting except the alignment weight $\beta$.

\paragraph{Offline training.}
The ARC-AGI-1 models train from scratch on the 401{,}589-record
corpus. The alignment term uses the input canvas as trajectory state
$T_0$ and is active for the whole run under the warmup schedule
$\beta_e=\beta\min(1,e/5)$ with the epoch index $e$ starting at one;
untraced instances contribute the final-state term alone. Training
runs data-parallel over the eight GPUs with a static graph.

\paragraph{ARC-AGI-2 warm start.}
The ARC-AGI-2 models start from the corresponding ARC-AGI-1 EMA
checkpoint and are trained for a further 100 epochs on the
882{,}270-record ARC-AGI-2 corpus under the same schedule. Only the
weights transfer: the per-task token table is re-initialized for the
new task set, and the optimizer state and epoch counter are reset, so
the run is a fresh cosine cycle rather than a continuation.

\paragraph{Test-time training.}
Following VARC \cite{hu2025arcvisionproblem}, each evaluation task is
augmented into 51 variants: the original task, plus five geometric
views (rotations by $90^{\circ}$, $180^{\circ}$, $270^{\circ}$ and the
two axis flips), each in its original colors and under nine random
color permutations. All variants' demonstration pairs form the
training set of that task, each variant carrying its own task
identity, and the per-task token table is re-initialized to that
number of entries. Test-time training optimizes the plain
cross-entropy of the final canvas only: no chain supervision and no
change weighting are used at test time, so the transformation chains
are purely an offline resource and every compared system runs the
identical test-time objective.

\onecolumn
\section{Additional Process Visualizations}
\label{app:visualization}

Figure~\ref{fig:app-execute} complements the three examples in the
main paper with three additional tasks, showing how predictions and
workspace assignments evolve across loop iterations.

\begin{center}
\centering
\includegraphics[
  width=\textwidth,
  height=0.74\textheight,
  keepaspectratio,
  trim=4 12 2 7,
  clip
]{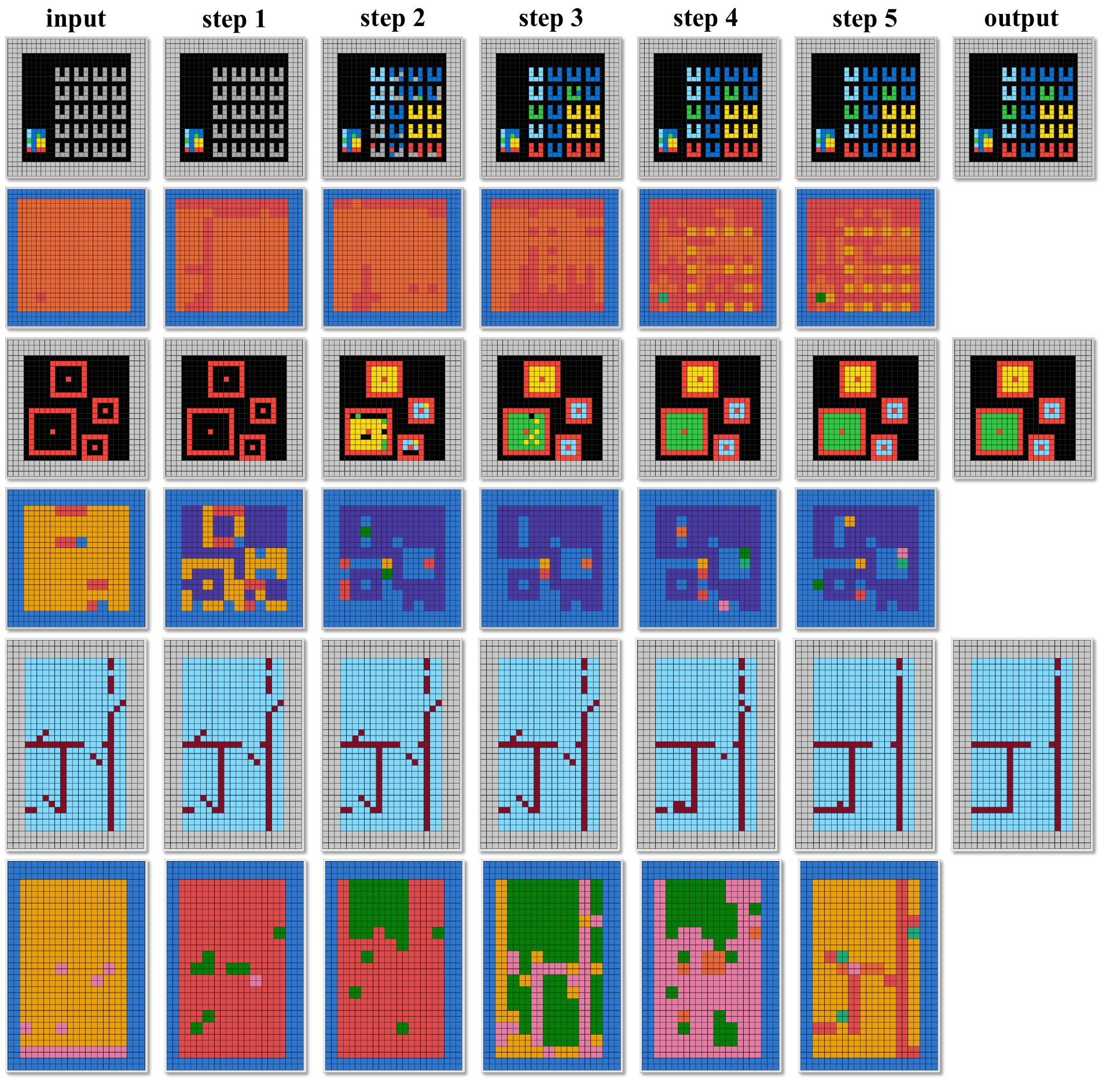}
\captionof{figure}{\textbf{Additional per-step predictions and slot assignments}
on three tasks. Prediction rows show the input and six loop outputs.
Assignment rows show the argmax over the eight workspace slots at each
iteration.}
\label{fig:app-execute}
\end{center}

\medskip
\noindent
The remaining pages show representative training records from
ARC-AGI-1 and ARC-AGI-2 ARC-GEN synthesis and from verifier-traced
RE-ARC data. Each record is rendered as an input, its intermediate
transformation states, and the final output.

\includepdf[
  pages=-,
  fitpaper=true,
  pagecommand={\thispagestyle{empty}},
  pagecommand*={%
    \refstepcounter{section}%
    \label{app:record-gallery}%
    \pdfbookmark[1]{F Training-Record Gallery}{app-gallery-bkm}%
  },
  picturecommand*={%
    \put(30,739){\color{white}\rule{552pt}{43pt}}%
    \put(0,758){%
      \makebox(612,0){\large\bfseries
        \thesection\quad Training-Record Gallery}%
    }%
  }
]{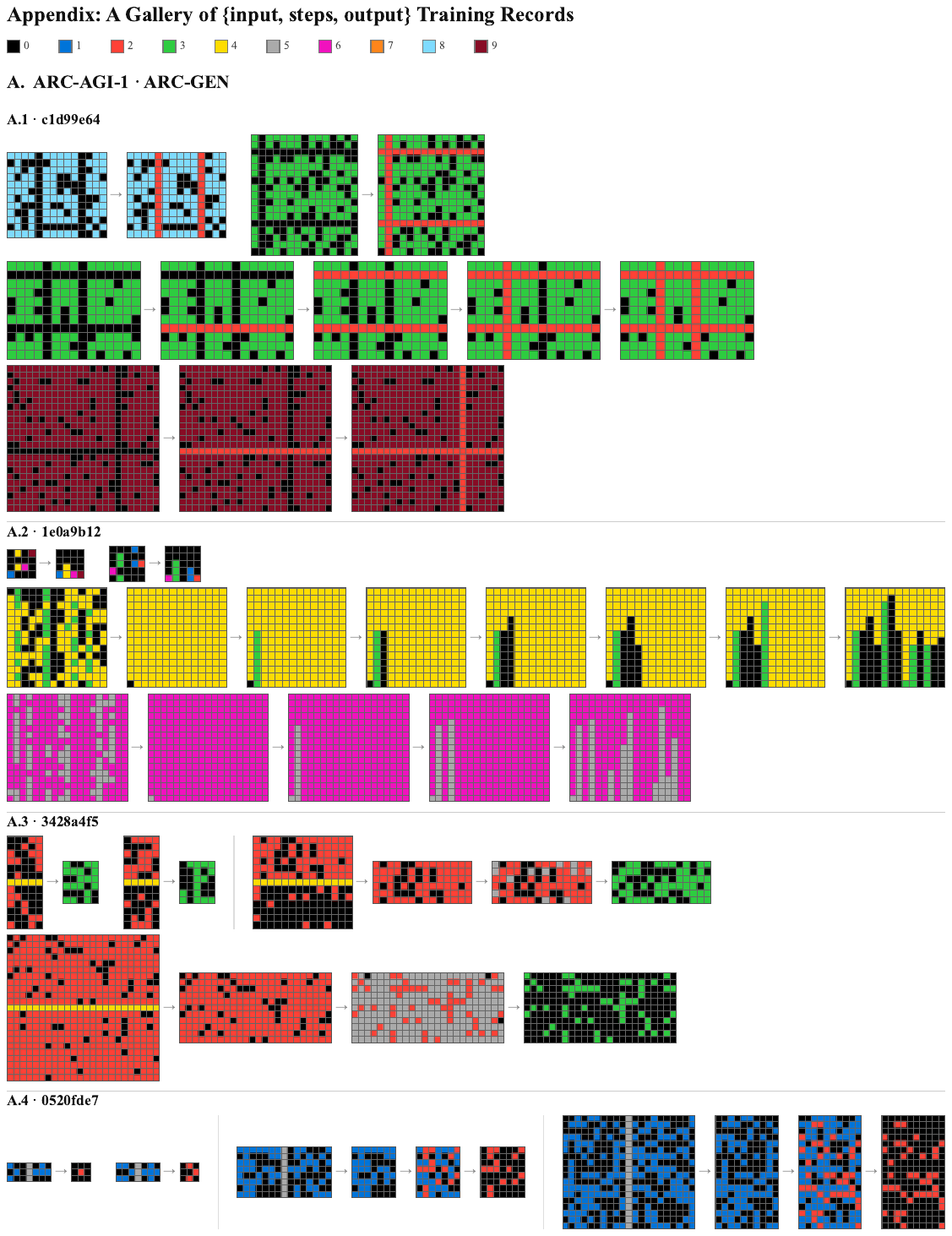}

\end{document}